\documentclass[sigconf]{acmart}

\usepackage{booktabs} 

\setcopyright{rightsretained}

\acmConference[]{Archive}{Submission ID}{2286011}
\copyrightyear{2018}

\begin{document}
\title{EasyConvPooling: Random Pooling with Easy Convolution for Accelerating Training and Testing}

\author{Jianzhong Sheng}
\authornote{Author}
\affiliation{%
  \institution{Huazhong University of Science and Technology\\City University of Hong Kong}
}
\email{csjianzhong@gmail.com}

\author{Chuanbo Chen}
\affiliation{%
  \institution{Huazhong University of Science and Technology}
}
\email{chuanboc@163.com}

\author{Chenchen Fu}
\affiliation{%
  \institution{City University of Hong Kong}
}
\email{chencfu2@cityu.edu.hk}

\author{Chun Jason Xue}
\authornote{Corresponding Author}
\affiliation{%
  \institution{City University of Hong Kong}
  }
\email{jasonxue@cityu.edu.hk}


\begin{abstract}
Convolution operations dominate the overall execution time of Convolutional Neural Networks (CNNs). This paper proposes an easy yet efficient technique for both Convolutional Neural Network training and testing. The conventional convolution and pooling operations are replaced by Easy Convolution and Random Pooling (ECP). In ECP, we randomly select one pixel out of four and only conduct convolution operations of the selected pixel. As a result, only a quarter of the conventional convolution computations are needed. Experiments demonstrate that the proposed EasyConvPooling can achieve 1.45x speedup on training time and 1.64x on testing time. What's more, a speedup of 5.09x on pure Easy Convolution operations is obtained compared to conventional convolution operations.
\end{abstract}

%
%



\maketitle

\section{Introduction}
Convolutional Neural Networks (CNNs) are a promising class of machine learning algorithms that achieves remarkable performance in various computer vision tasks, e.g., image classification \cite{krizhevsky2012imagenet}. One of the key reason for this success is their deep architecture \cite{montufar2014number}. It has been proved that deeper architecture makes better performance. As a result, the performance of CNNs over past few years has been improved mainly by designing a deeper architecture. It is not uncommon for a neural network to have massive parameters in its model, costing more time to train and test the network.

In this study, we propose an effective technique called EasyConvPooling (ECP) to accelerate both training and testing. EasyConvPooling is consist of two parts: Easy Convolution and Random Pooling. In Random Pooling, we select one pixel out of four randomly, and then compute convolution of the selected pixel only. This leads to reduction in 75\% convolution computation compared to conventional convolution and thus reducing both training time and testing time.

In order to realize the proposed method, we are facing two questions. The first question is how to determine the selected pixel in Random Pooling and how to obtain its index for conducting Easy Convolution in the upper layer. For selecting pixel, we randomly appoint one pixel out of four to be the ``lucky'' pixel; for its index, we keep the index of the ``lucky'' pixel which is appointed before pooling. This does not lead to significant loss in accuracy.

The second question is how to conduct Easy Convolution in selected mode and keep the shape of output feature map unchanged. In order to solve this problem, we determine the mode of Easy Convolution according to the index of the selected pixel to assure that they match. Based on the experiments, we find that conducting Random Pooling alone does reduce training time and testing time. The reduction grows to be more significant when combined with Easy Convolution. Experimental results demonstrate that the proposed ECP achieves 1.45x speedup on training time and 1.64x speedup on testing time. In addition, we obtain a speedup of 5.09x on pure Easy Convolution operations compared to conventional convolution operations.

The contributions of this work are as follows:
\begin{itemize}
\item Proposes a novel EasyConvPooling technique to conduct convolution and pooling, in which only 25\% of convention convolution operations is needed.
\item Proposes an universal technique to accelerate both training and testing.
\item The proposed novel technique (ECP) can be transfered to any other platform supporting Python.
\end{itemize}

Remainder of this paper is organized as follows.
Section 2 summarizes the related work.
Section 3 presents the proposed method.
Section 4 compares the proposed method with the state-of-the-art schemes. 
Finally, we conclude the paper in Section 5.

\begin{figure*}[t]
\centering
\includegraphics[scale=0.6]{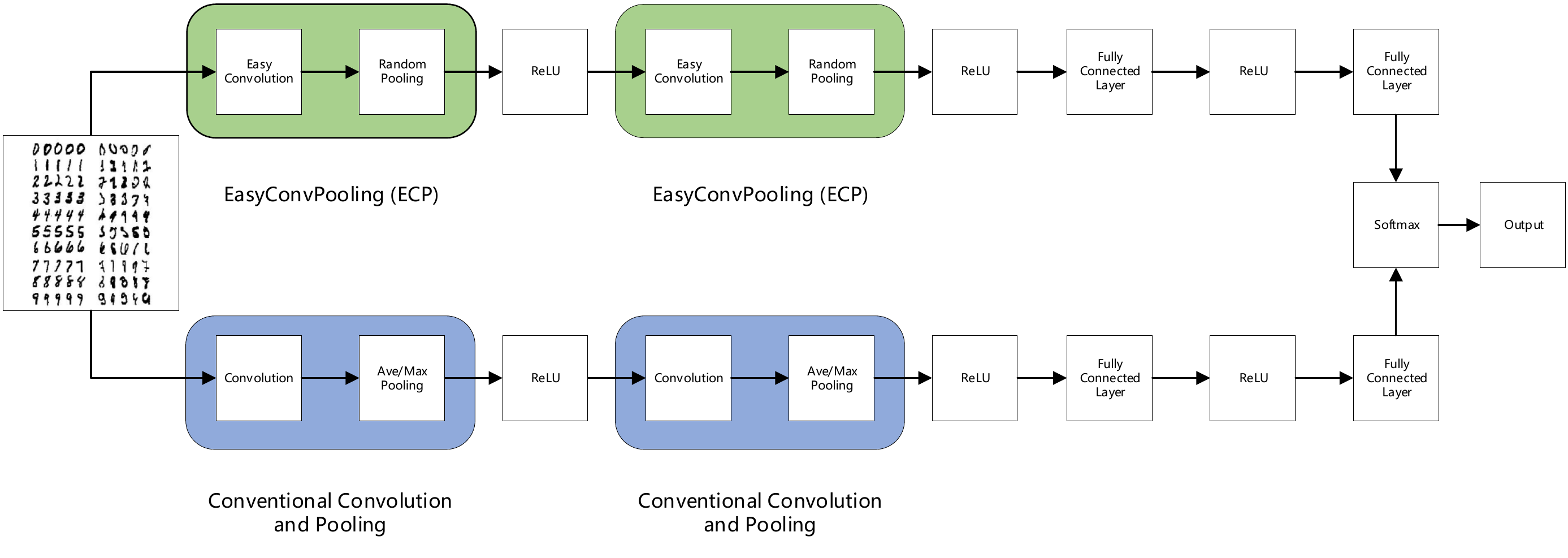}
\caption{Architecture of the Network.}
\label{fig:1}
\end{figure*}

\section{Related Work}
\subsection{Accelerating Convolutional Neural Network}
Lots of algorithm are proposed for accelerating convolutional neural network. Han et al. \cite{han2015learning,han2015deep,han2016eie,han2017ese,han2016dsd} proposed pruning methods to cut off unimportant connections in fully connected layer to avoid useless computation. These methods can be applied to CPU, GPU, FPGA and ASIC, achieving speedup of 13x, 10x for VGG-16 \cite{simonyan2014very} and LSTM \cite{hochreiter1997long} respectively. These optimizations on fully connected layers differ from ours, we are focusing on the most time consuming part of convolutional neural network.

Liu et al. \cite{liu2015sparse}, Yuan et al. \cite{yuan2006model}, Feng et al. \cite{feng2015learning}, Lebedev et al. \cite{lebedev2016fast}, Wen et al. \cite{wen2016learning}, Denil et al. \cite{denil2013predicting}, Denton et al. \cite{denton2014exploiting}, Jaderberg et al. \cite{jaderberg2014speeding}, Ioannou et al. \cite{ioannou2015training} and Tai et al. \cite{tai2015convolutional} proposed weight sparsity methods to utilize the sparsity in weights. They increased zero elements in weight matrices to make the matrices sparse, and thus reduced data to be stored and computed. Their methods make use of sparsity to reduce computation, however, it remains unknown how much time they cost to transfer these weights to be sparse.

Courbariaux et al. \cite{courbariaux2015binaryconnect,courbariaux2016binarized}, Lin et al. \cite{lin2013network}, Baldassi et al. \cite{baldassi2015subdominant}, Cheng et al. \cite{cheng2015training}, Kim et al. \cite{kim2016bitwise} and Hwang et al. \cite{hwang2014fixed} proposed Binary and Ternary network to make weights constrained to $0$ and $\pm 1$. As a result, lots of multiplications are reduced, this makes it possible for FPGA. Their methods convert weights to $0$ and $\pm 1$, and make use of $0$ and $\pm 1$ to reduce computation while we just throw extra computation away directly!

MeProp was proposed by Sun et al. \cite{sun2017meprop} in 2017. They made optimization in back propagation and achieved amazing performance on back propagation time. However, their technique can only speed up the back propagation and do nothing to the forward path, making no contribution to testing time. In the proposed ECP, we can speed up both training and testing. 
\subsection{Pooling Algorithms}
Max Pooling \cite{yang2009linear} and Average Pooling \cite{lecun1998gradient} are mostly used pooling methods in conventional Convolutional Neural Networks. Max Pooling conducts pooling by selecting the max value pixel to represent the output of pooling window while Average Pooling computes their average value to be output. Both Max Pooling and Average Pooling make full convolution operations of the four pixels required in pooling window to output one pixel only. This is where we can reduce 75\% convolution operations by conducting Easy Convolution and Random Pooling.

\section{Proposed ECP}
We propose an easy yet efficient technique called EasyConvPooling (ECP) for Convolutional Neural Networks to conduct convolution operations and pooling. In the proposed method ECP, only 25\% of original convolution operations are done, which reduces 75\% multiplications in convolutions with little loss in accuracy. ECP is consist of two parts: Easy Convolution and Random Pooling. Here is how to conduct ECP:
\begin{itemize}
\item Randomly set Mode K.
\item Determine the positions of selected pixels for Random Pooling and Easy Convolution.
\item Conduct Easy Convolution on the selected pixels and pad the neighbor pixels to recover the output shape for pooling layer.
\item Conduct Random Pooling.
\end{itemize}

In the following subsection, we first present the architecture of the network, conventional convolution and pooling, then describe Random Pooling and Easy Convolution in details.

\subsection{Architecture}

Figure \ref{fig:1} shows overall architecture of the proposed network for conducting ECP compared to conventional convolution and pooling. In Figure \ref{fig:1}, we design a two-convolution neural network with two fully connected layers. Each convolution layer is followed by a pooling layer and a ReLU layer. In the fully connected layers, we add one ReLU layer at the end of the first layer and connect the second fully connected layer to the Softmax layer directly.

The upper part of Figure \ref{fig:1} is the proposed ECP technique and the lower part is conventional way to do convolution and pooling, such as Average Pooling and Max Pooling. In ECP, we replace conventional convolution and pooling operations by Easy Convolution and Random Pooling. Both Easy Convolution and Random Pooling have a Mode K to control operation mode. In order to assure that they are matched under the same Mode K, each Easy Convolution layer is followed by a Random Pooling layer.

\begin{figure*}[ht]
\centering
\includegraphics[scale=0.5]{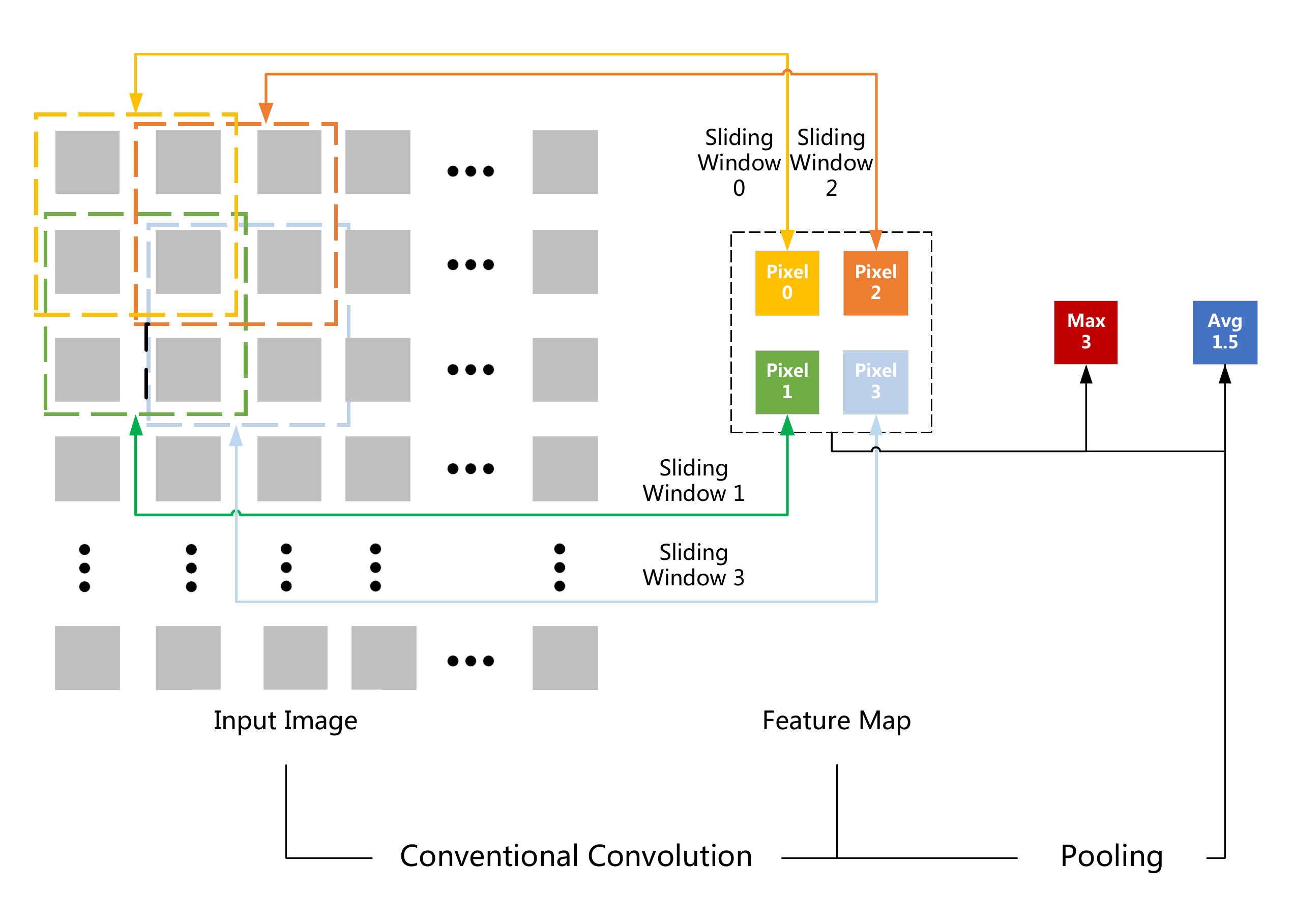}
\caption{Conventional Convolution and Pooling.}
\label{fig:2}
\end{figure*}

\subsection{Conventional Convolution and Pooling}
Convolution operations occupy the most time of CNNs, and in Figure \ref{fig:2} we look into conventional convolution and pooling to make an overall view of the conventional convolution and pooling. In the following subsection, we will describe and compare conventional convolution and pooling with the proposed Easy Convolution and Random Pooling in details.

In Figure \ref{fig:2}, every sliding (convolution) window is consist of four pixels, taking kernel size $2\times2$ for easy demonstration, and every time we make convolution of a sliding window (weights) and input pixels beneath it to form a feature map element. Considering one step stride, sliding windows are overlapped. After computing one feature map element out, we move the sliding window one step right to compute another feature map element and as well as in the second row. Finally, we achieve a feature map for pooling. Here is how we compute convolution:
\begin{equation*}
W*x(m,n)=\sum_{u}\sum_{v}W(u,v)x(m+u,n+v)
\end{equation*}
where $x$ is an input image and $W$ is a weight matrix of the convolution filter. The operator `$*$' means 2D convolution.

In the pooling layer, output is calculated by selecting one pixel out of four to represent the whole four pixels. In Average Pooling, the output is the average value of the four pixels in feature map; in Max Pooling, we select the max value pixel as the output pixel.

In short, we compute four conventional convolutions to form the pooling elements required for pooling window. However, the output of both Average Pooling and Max Pooling are one pixel only, wasting extra 75\% convolutions. If we can determine which pixel to be selected in the pooling layer, we can reduce the extra 75\% convolutions in convolution layer. That's where we benefit in the proposed Easy Convolution and Random Pooling.

\subsection{Random Pooling}
In conventional convolution, we need to calculate the outputs of every convolution window to make the feature map, however, in the pooling layer, only one pixel out of four (stride = 2) is chosen to represent the output of the pooling window. In Average Pooling, we compute the average of the four pixels to make the output of pooling window and in Max Pooling we make the output by choosing the max value pixel. Random Pooling just randomly select one pixel out of four to represent the output of pooling window reducing 75\% extra convolutions.

In Random Pooling, we obtain the index of the selected pixel by setting Random Pooling Mode K. Random Pooling Mode K stands for the position index of the four pixels in the pooling windows. It varies from 0 to 3, from up to down and left to right. The Random Pooling Mode K is randomly set before pooling so that we can figure out how to conduct Easy Convolution in the upper layer.
\begin{figure*}[ht]
\centering
\includegraphics[scale=0.5]{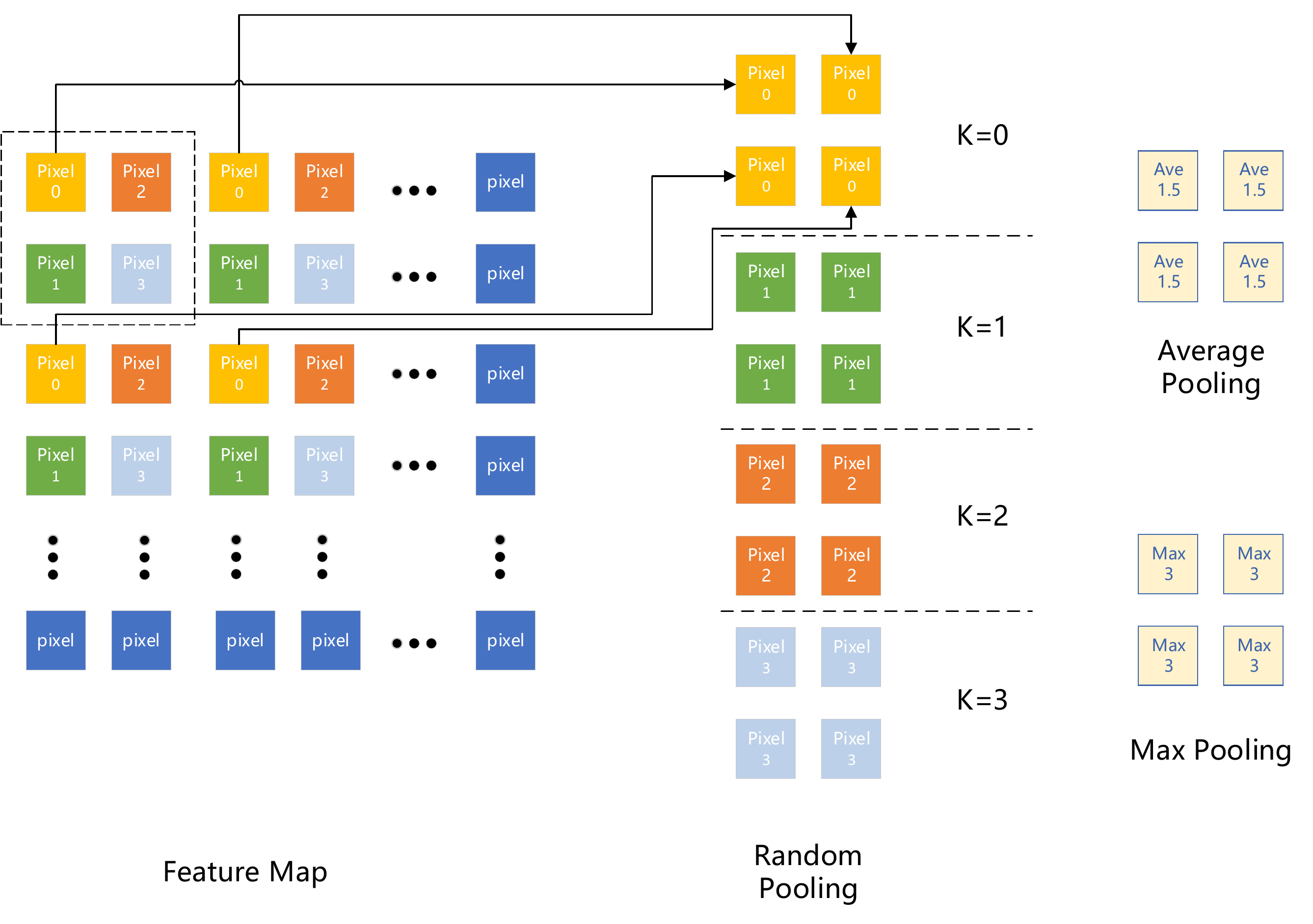}
\caption{Random Pooling vs Average/ Max Pooling.}
\label{fig:3}
\end{figure*}

Figure \ref{fig:3} demonstrates how exactly Random Pooling works. Once the Random Pooling Mode K is set, we can determine which pixel to be selected in the dotted pooling window. Mode 0 means pixel 0 is selected from the dotted pooling window every time. After selecting the first pixel 0 element, we slide the pooling window two step right to obtain the second pixel 0 element. The pooling window slides from left to right, top to down with two strides every time. Finally, the output feature map of Random Pooling is formed by those pixel 0 elements. In Mode 1, 2 and 3, the same operations are done to pixel 1, 2 and 3 elements. Random Pooling actually always select the pixel of the same position in the pooling windows to make up the outputs of the pooling windows and thus form the output feature map of the pooling layer. Various Mode K means various pixel position in the pooling window.

In conventional Average Pooling/ Max Pooling, the output of pooling window is always the Averaged/ Maxed value of the pooling window. In the middle of Figure \ref{fig:3} is the proposed Random Pooling, and beside it is conventional Average/ Max Pooling.

\subsection{Easy Convolution}
In order to match and control the Easy Convolution Mode with Random Pooling Mode, we use the same K to control Easy Convolution Mode. Due to the overlapping in convolution sliding window, we need two variables to locate the position of selected convolution window. In programming level, we make the two variables in Easy Convolution to match the Random Pooling Mode K so that we can use the same K in Easy Convolution layer to extract the same position elements matching Random Pooling. 
\begin{figure*}[ht]
\centering
\includegraphics[scale=0.5]{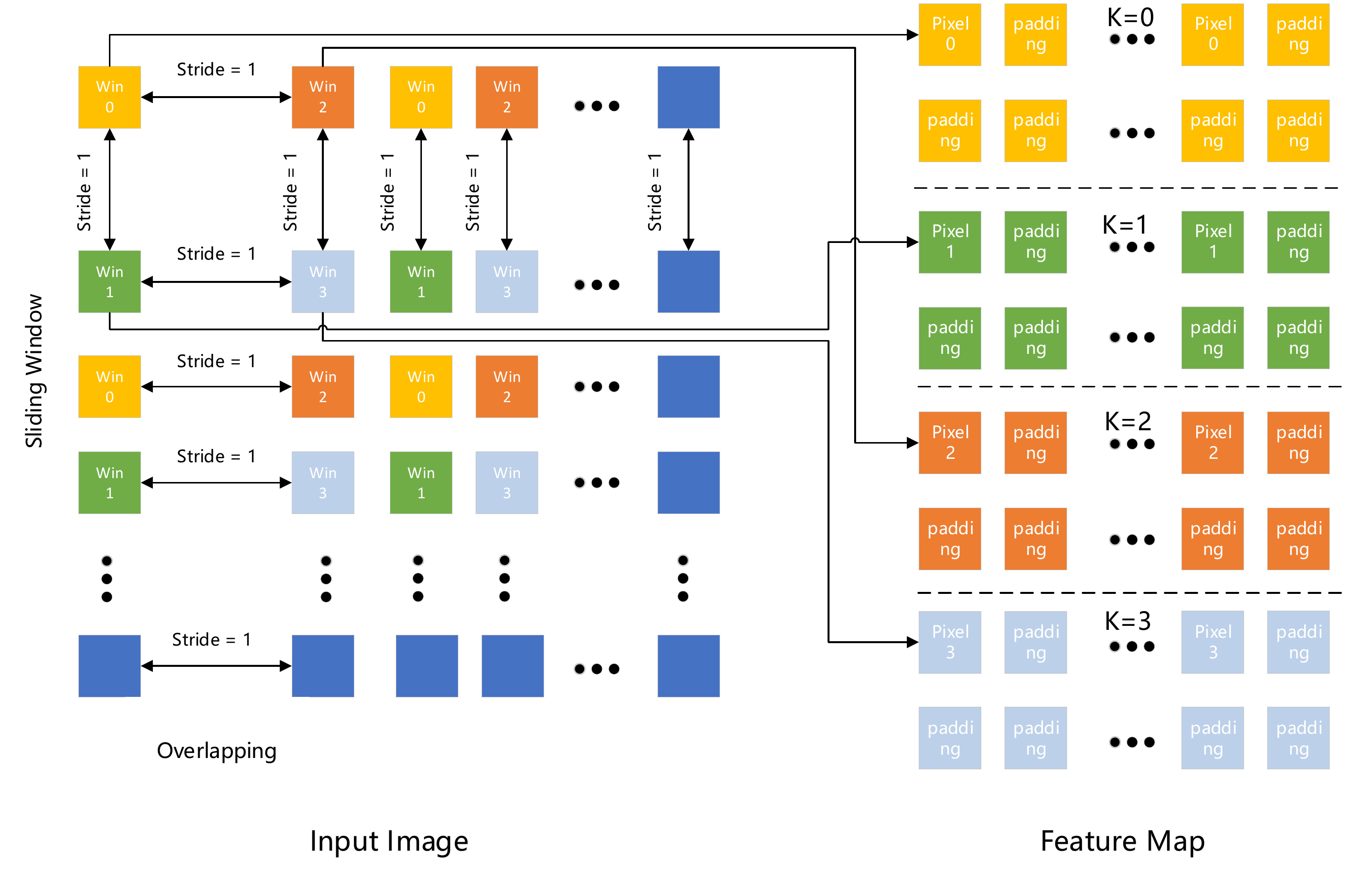}
\caption{Easy Convolution.}
\label{fig:4}
\end{figure*}

In subsection Random Pooling, we obtain the selected pixel's index by setting Random Pooling Mode K and then we conduct Easy Convolution using the same Mode K. Figure \ref{fig:4} shows how we conduct the Easy Convolution operations, it's very similar to Random Pooling.

In Figure \ref{4}, every convolution window on input image contains several weights. In order to obtain feature map for the next layer, convolution operations are carried out on these input image with sliding convolution window. Window 0 is the selected window for producing selected pixel 0 element for pooling window in pooling layer. Every window 0 is a sliding convolution window over input image under Mode 0. The output of these convolution window is pixels needed in pooling layer for Random Pooling. The first window 0 produces the first pixel 0 element in the pooling window in Figure \ref{fig:3}, and the second window 0 produces the second pixel 0 element in the pooling window. The convolution window slides over the input image to produce the output feature of convolution layer. Various Mode K determine various window K to produce various pixel K element needed in the pooling window. After sliding from left to right and up to down, the feature map of convolution layer is formed. 

In conventional convolution, sliding convolution window slides over all input image area to produce feature map elements. In Easy Convolution, sliding convolution window slides only to window K position to extract selected data, reducing 75\% extra data with a quarter of original shape.

After extracting data we need from input image, we can easily compute convolution as usual, reducing 75\% convolutions. The remaining problem is how we can get the pruned shape back. In some situation, we make use of padding technique to keep the output shape unchanged. We add padding to the output of Easy Convolution to restore the shape of the feature map so that the network can run as usual. For Easy Convolution, we pad the same value to its neighbor empty pixels as shown in Figure \ref{fig:4}.

\section{Experiments and Evaluations}
To demonstrate that the proposed technique ECP is effective and reliable, we perform experiments on various hidden layers compared with Average Pooling and Max Pooling under different Mode K. We coded a one-convolution layer network and a two-convolution layer network to evaluate ECP's performance under various depth of layers. All the codes are written in Python without any framework, and all the experiments are conducted on CPU, making it universal to all platform supporting Python.
\subsection{Experimental Settings}
We set batch size to 50 and learning rate to 0.001 in the experiments. For MNIST \cite{lecun1998gradient}, the input dimension is 28$\times$28 =784, and the output dimension is 10. The experiments are conducted on Intel(R) Core i7-7700HQ 2.80GHz CPU with Python 3.6.3 installed on Windows 10 operation system.

Table \ref{1} and Table \ref{2} show the parameters of the networks under various hidden layers.

\begin{table}[]
\centering
\caption{Network Parameters of One-convolution layer CNN.}
\small
\label{1}
\begin{tabular}{|l|l|}
\hline
\begin{tabular}[c]{@{}l@{}}Layer   Name\end{tabular} & Parameter                                                           \\ \hline
Input image                                            & \begin{tabular}[c]{@{}l@{}}size: 28$\times$28, channel:\\   1\end{tabular} \\
Convolution                                            & kernel: 5$\times$5, channel: 20                                            \\
Pooling                                                & kernel: 2$\times$2, stride: 2                                              \\
ReLU                                                   &                                                                     \\
Fully connected                                        & channel: 100                                                        \\
ReLU                                                   &                                                                     \\
Fully connected                                        & channel: 10                                                         \\
Softmax                                                &                                                                     \\ \hline
\end{tabular}
\end{table}

\begin{table}[]
\centering
\caption{Network Parameters of Two-convolution layer CNN.}
\small
\label{2}
\begin{tabular}{|l|l|}
\hline
\begin{tabular}[c]{@{}l@{}}Layer   Name\end{tabular} & Parameter                                                           \\ \hline
Input image                                            & \begin{tabular}[c]{@{}l@{}}size: 28$\times$28, channel:\\   1\end{tabular} \\
Convolution                                            & kernel: 5$\times$5, channel: 20                                            \\
Pooling                                                & kernel: 2$\times$2, stride: 2                                              \\
ReLU                                                   &                                                                     \\
Convolution                                            & kernel: 5$\times$5, channel: 32                                            \\
Pooling                                                & kernel: 2$\times$2, stride: 2                                              \\
ReLU                                                   &                                                                     \\
Fully connected                                        & channel: 100                                                        \\
ReLU                                                   &                                                                     \\
Fully connected                                        & channel: 10                                                         \\
Softmax                                                &                                                                     \\ \hline
\end{tabular}
\end{table}

\subsection{Experimental Results on Time Performance}
In order to verify the time performance of the proposed technique ECP, we evaluate both training time and testing time at the same time. Furthermore, we design a special test on pure convolution with the proposed ECP and conventional convolution method to compare their real operation time on convolution. This special test is conducted on MNIST database for 100 epochs, and we have tested it for several times.

Table \ref{3} shows the overall time performance of the proposed ECP compared with the conventional Max/ Average Pooling. The first column shows the exact epoch when testing accuracy first reaches 98\%, and the others indicate time performance. After applying ECP, we achieve 1.45x speedup on training time and 1.64x on testing time compared to Average Pooling. In terms of the pure convolution time, we achieve the speedup of 5.09x. This speedup is even larger than the theoretical speedup value. This is because of the limitation of the memory space. Less convolution data can save the space of memory and thus avoid the content switch operations due to the lack of memory. In addition, based on the result over Iteration, we can be sure that the ECP technique does not lead to more training data.

\begin{table*}
\centering
\caption{Overall Time Performance of ECP vs Max/ Average Pooling.}
\label{3}
\begin{tabular}{|l|l|l|l|l|}
\hline
Method      & Iter (98\%) & Training Time (ms)         & Testing Time (ms)         & PureConvTime (ms)       \\ \hline
Max Pooling & 3           & 299650.15                  & 23814.87                  & 4496.77                 \\ \hline
ECP         & 4           & \textbf{243077.47 (1.23x)} & \textbf{13084.98 (1.82x)} & \textbf{883.65 (5.09x)} \\ \hline
Method      & Iter (98\%) & Training Time (ms)         & Testing Time (ms)         & PureConvTime (ms)       \\ \hline
Ave Pooling & 5           & 352544.96                  & 21426.24                  & 4496.77                 \\ \hline
ECP         & 4           & \textbf{243077.47 (1.45x)} & \textbf{13084.98 (1.64x)} & \textbf{883.65 (5.09x)} \\ \hline
\end{tabular}
\end{table*}

\subsection{Experimental Results on Accuracy}
Besides the time performance, accuracy is another critical parameter in both training and testing steps. Considering the randomness of ECP, it may lead to drop in accuracy. In order to figure out this, we first conduct experiments by training the MNIST dataset for 200 times to make sure we can get its best accuracy during experiment. Results indicate that 100 epochs are already enough. For most situation, they achieve their best accuracy within 80 epochs. Sometimes we get worse accuracy while training more due to overfitting. So in the experiments, we decide to evaluate the accuracy within 100 epochs, which is more valuable than training another more 100 epochs to gain accuracy improvement less than 0.5\%.

The experiments are conducted on a two-convolution layer network demonstrated in Figure \ref{fig:1}. Results in Table \ref{4} indicate the proposed ECP achieves good improvement with little loss in accuracy.

The experiment does not only consider the accuracy performance of ECP compared with Max Pooling and Average Pooling but also takes Mode K into consideration to evaluate the Robustness of the proposed ECP. The results are reliable, and Mode K will be discussed in detail in the following subsection.
\begin{table*}[]
\centering
\caption{Random Pooling and ECP vs Max /Average Pooling under Mode K.}
\label{4}
\begin{tabular}{|l|l|l|l|l|}
\hline
Method      & Iter (98\%) & Training Time (ms)         & Testing Time (ms)         & Best Accuraacy (\%)    \\ \hline
Max Pooling & 3           & 299650.15                  & 23814.87                  & 99.17                  \\ \hline
Random k=0  & 4           & \textbf{280866.26 (1.07x)} & \textbf{20691.98 (1.15x)} & \textbf{98.65 (-0.52)} \\ \hline
Random k=1  & 4           & \textbf{281443.49 (1.06x)} & \textbf{20689.19 (1.15x)} & \textbf{98.78 (-0.39)} \\ \hline
Random k=2  & 5           & \textbf{277848.33 (1.08x)} & \textbf{20422.44 (1.17x)} & \textbf{98.77 (-0.40)} \\ \hline
Random k=3  & 4           & \textbf{282693.13 (1.06x)} & \textbf{20577.25 (1.16x)} & \textbf{98.78 (-0.39)} \\ \hline
ECP k=0     & 4           & \textbf{243077.47 (1.23x)} & \textbf{13084.98 (1.82x)} & \textbf{98.81 (-0.36)} \\ \hline
ECP k=1     & 5           & \textbf{243607.47 (1.23x)} & \textbf{13403.70 (1.78x)} & \textbf{98.67 (-0.50)} \\ \hline
ECP k=2     & 5           & \textbf{240106.84 (1.25x)} & \textbf{13271.43 (1.80x)} & \textbf{98.65 (-0.52)} \\ \hline
ECP k=3     & 5           & \textbf{243466.94 (1.23x)} & \textbf{13303.02 (1.80x)} & \textbf{98.77 (-0.40)} \\ \hline
Method      & Iter (98\%) & Training Time (ms)         & Testing Time (ms)         & Best Accuraacy (\%)    \\ \hline
Ave Pooling & 5           & 352544.96                  & 21426.24                  & 98.69                  \\ \hline
Random k=0  & 4           & \textbf{280866.26 (1.26x)} & \textbf{20691.98 (1.04x)} & \textbf{98.65 (-0.04)} \\ \hline
Random k=1  & 4           & \textbf{281443.49 (1.25x)} & \textbf{20689.19 (1.04x)} & \textbf{98.78 (+0.09)} \\ \hline
Random k=2  & 5           & \textbf{277848.33 (1.27x)} & \textbf{20422.44 (1.05x)} & \textbf{98.77 (+0.08)} \\ \hline
Random k=3  & 4           & \textbf{282693.13 (1.25x)} & \textbf{20577.25 (1.04x)} & \textbf{98.78 (+0.09)} \\ \hline
ECP k=0     & 4           & \textbf{243077.47 (1.45x)} & \textbf{13084.98 (1.64x)} & \textbf{98.81 (+0.12)} \\ \hline
ECP k=1     & 5           & \textbf{243607.47 (1.45x)} & \textbf{13403.70 (1.60x)} & \textbf{98.67 (-0.02)} \\ \hline
ECP k=2     & 5           & \textbf{240106.84 (1.47x)} & \textbf{13271.43 (1.61x)} & \textbf{98.65 (-0.04)} \\ \hline
ECP k=3     & 5           & \textbf{243466.94 (1.45x)} & \textbf{13303.02 (1.61x)} & \textbf{98.77 (+0.08)} \\ \hline
\end{tabular}
\end{table*}
\subsection{Varying Mode K}
Another interesting problem is to check the role of Mode K. In Random Pooling and Easy Convolution, we randomly set a parameter by Mode K, and it determines how to conduct Random Pooling and where to apply the Easy Convolution.
In the following experiment, we test what happens if we vary the parameter Mode K.

\begin{figure}
\centering
\includegraphics[width=0.9\linewidth]{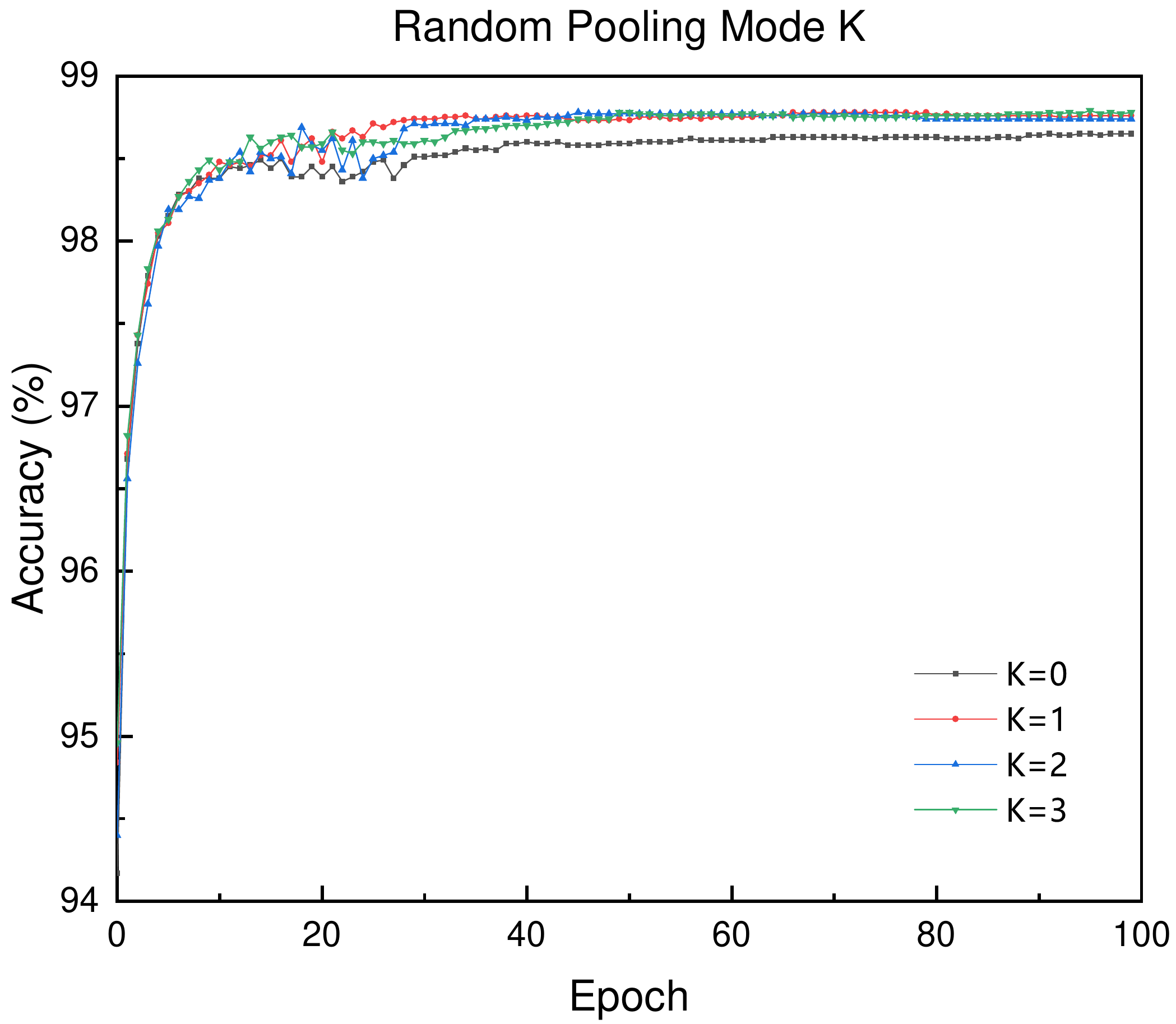}
\caption{Random Pooling Convergence under Mode K.}
\label{fig:5}
\end{figure}

\begin{figure}
\centering
\includegraphics[width=0.9\linewidth]{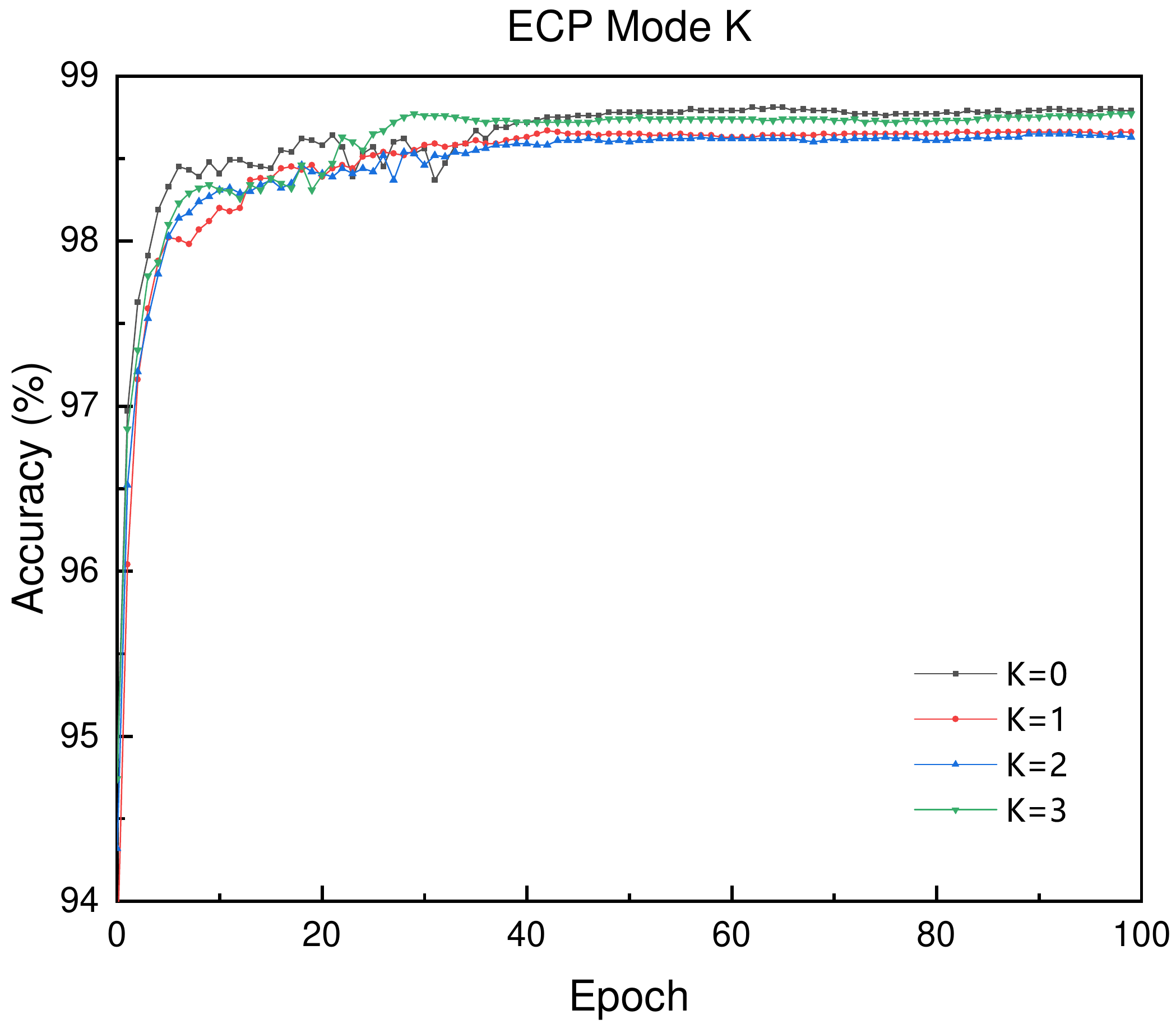}
\caption{ECP Convergence under Mode K.}
\label{fig:6}
\end{figure}

\begin{figure}
\centering
\includegraphics[width=0.9\linewidth]{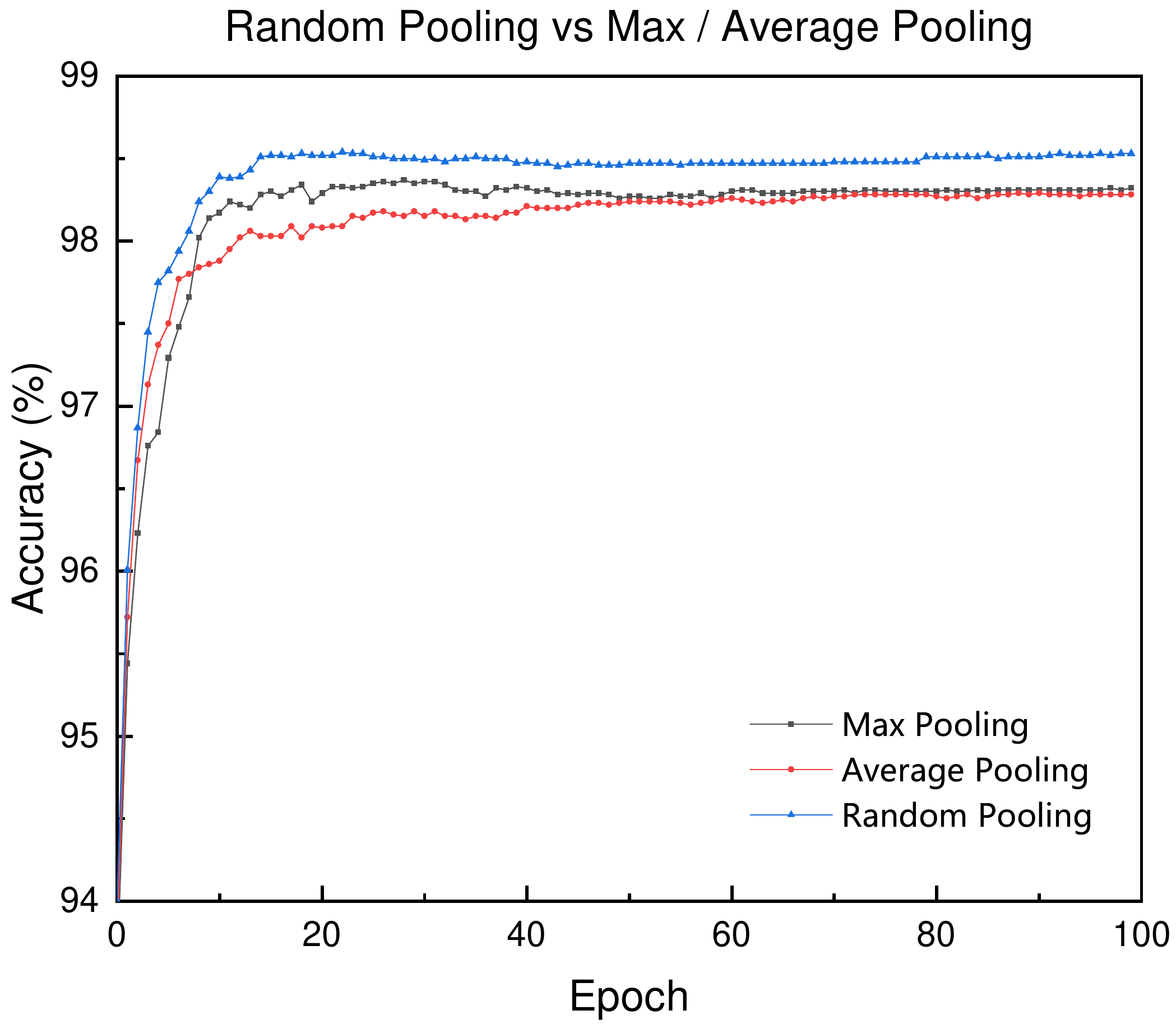}
\caption{Random Pooling Convergence vs Average/ Max Pooling under One-convolution Layer.}
\label{fig:7}
\end{figure}

To find out the role of Mode K plays in the proposed ECP, we design tests on a two-convolution network using Random Pooling technique only and ECP technique respectively. Figure \ref{fig:5} is Random Pooling under different Mode K, and Figure \ref{fig:6} shows results for ECP.

From the figures we can conclude that the randomly set Mode K is not crucial to the results, while it does affect the training process in some aspect. Randomly selected Mode K does not affect the overall convergence of the network no matter it's conducted alone or together with Easy Convolution, and Mode K has little influence on training time and testing time based on Table \ref{4}.

However, in Figure \ref{fig:5} and Figure \ref{fig:6}, the randomly selected Mode K seems to have some effect on the convergence in the very beginning of the training process and it seems to have effect on the final accuracy. But recall that the weights in the kernel are initiated randomly by uniform distribution. It can be noted that Mode K has little influence on ECP's time performance as well as accuracy.

\subsection{Varying Convolution Layers}
In this part, we evaluate the proposed ECP under various hidden layers: one-convolution network and 
two-convolution network. Table \ref{5} is the result of ECP compared with Max Pooling and Table \ref{6} is for Average Pooling.

From the tables, we notice that the performance compared to Average Pooling is better than that of Max Pooling. We gain more performance speedup compared to Average Pooling with a little accuracy improvement rather than drop. What's more, comparing to Max Pooling, the time performance of the proposed ECP is even better when we make the network deeper, with little accuracy loss.
\begin{table*}[]
\centering
\caption{ECP vs Max Pooling under Various Convolution Layers.}
\label{5}
\begin{tabular}{|l|l|l|l|l|l|}
\hline
ConvLayer & Method      & Iter (98\%) & Training Time (ms)         & Testing Time (ms)         & Best Accuracy (\%)     \\ \hline
1         & Max Pooling & 8           & 215387.59                  & 12500.73                  & 98.37                  \\ \cline{2-6} 
          & ECP         & 5           & \textbf{207065.46 (1.04x)} & \textbf{11520.13 (1.09x)} & \textbf{98.69 (+0.32)} \\ \hline
2         & Max Pooling & 3           & 299650.15                  & 23814.87                  & 99.17                  \\ \cline{2-6} 
          & ECP         & 4           & \textbf{243077.47 (1.23x)} & \textbf{13084.98 (1.82x)} & \textbf{98.81 (-0.36)} \\ \hline
\end{tabular}
\end{table*}

\begin{table*}[]
\centering
\caption{ECP vs Average Pooling under Various Convolution Layers.}
\label{6}
\begin{tabular}{|l|l|l|l|l|l|}
\hline
ConvLayer & Method      & Iter (98\%) & Training Time (ms)         & Testing Time (ms)         & Best Accuracy (\%)     \\ \hline
1         & Ave Pooling & 12          & 303580.05                  & 18579.89                  & 98.29                  \\ \cline{2-6} 
          & ECP         & 5           & \textbf{207065.46 (1.47x)} & \textbf{11520.13 (1.61x)} & \textbf{98.69 (+0.40)} \\ \hline
2         & Ave Pooling & 5           & 352544.96                  & 21426.24                  & 98.69                  \\ \cline{2-6} 
          & ECP         & 4           & \textbf{243077.47 (1.45x)} & \textbf{13084.98 (1.64x)} & \textbf{98.81 (+0.12)} \\ \hline
\end{tabular}
\end{table*}

\begin{figure}
\centering
\includegraphics[width=0.9\linewidth]{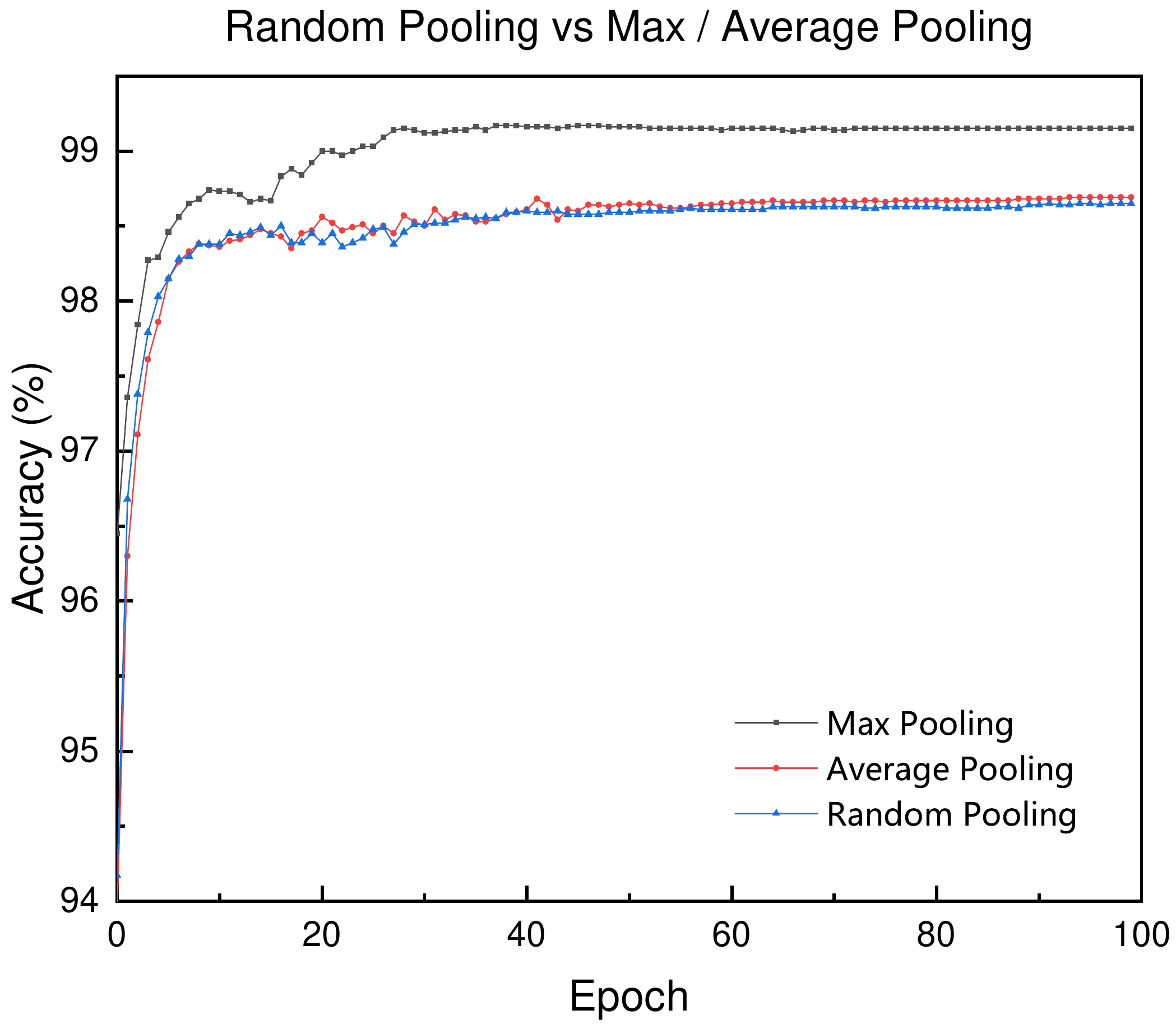}
\caption{Random Pooling Convergence vs Average/ Max Pooling under Two-convolution Layer.}
\label{fig:8}
\end{figure}

\subsection{Convergence under Various Convolution}
The proposed ECP is consist of two parts: Easy Convolution and Random Pooling. In order to compare convergence of Random Pooling alone with Average Pooling/ Max Pooling, we conduct the same conventional convolution in the upper convolution layer of Random Pooling. 

\begin{figure}
\centering
\includegraphics[width=0.9\linewidth]{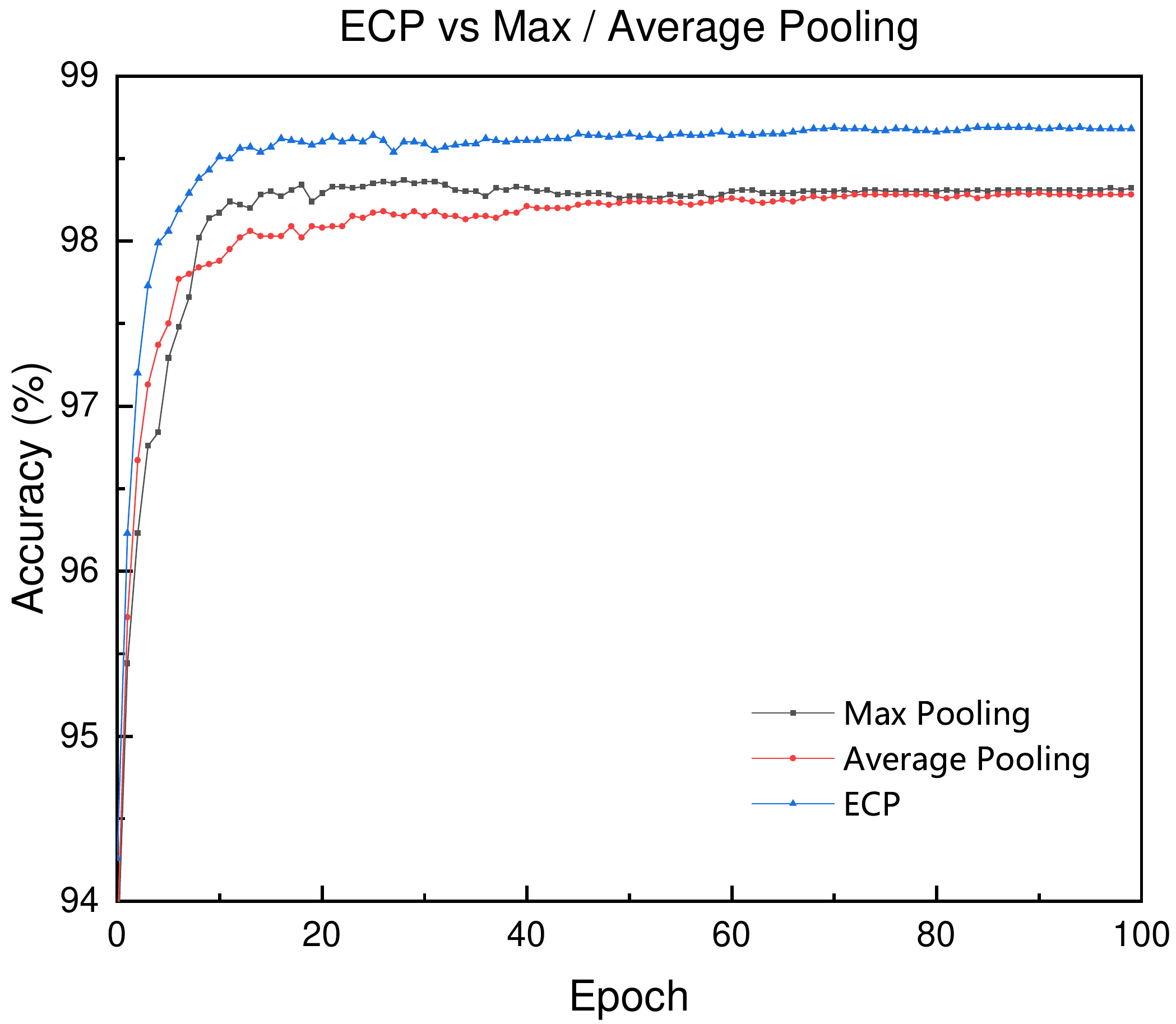}
\caption{ECP Convergence vs Average/ Max Pooling under One-convolution Layer.}
\label{fig:9}
\end{figure}

\begin{figure}
\centering
\includegraphics[width=0.9\linewidth]{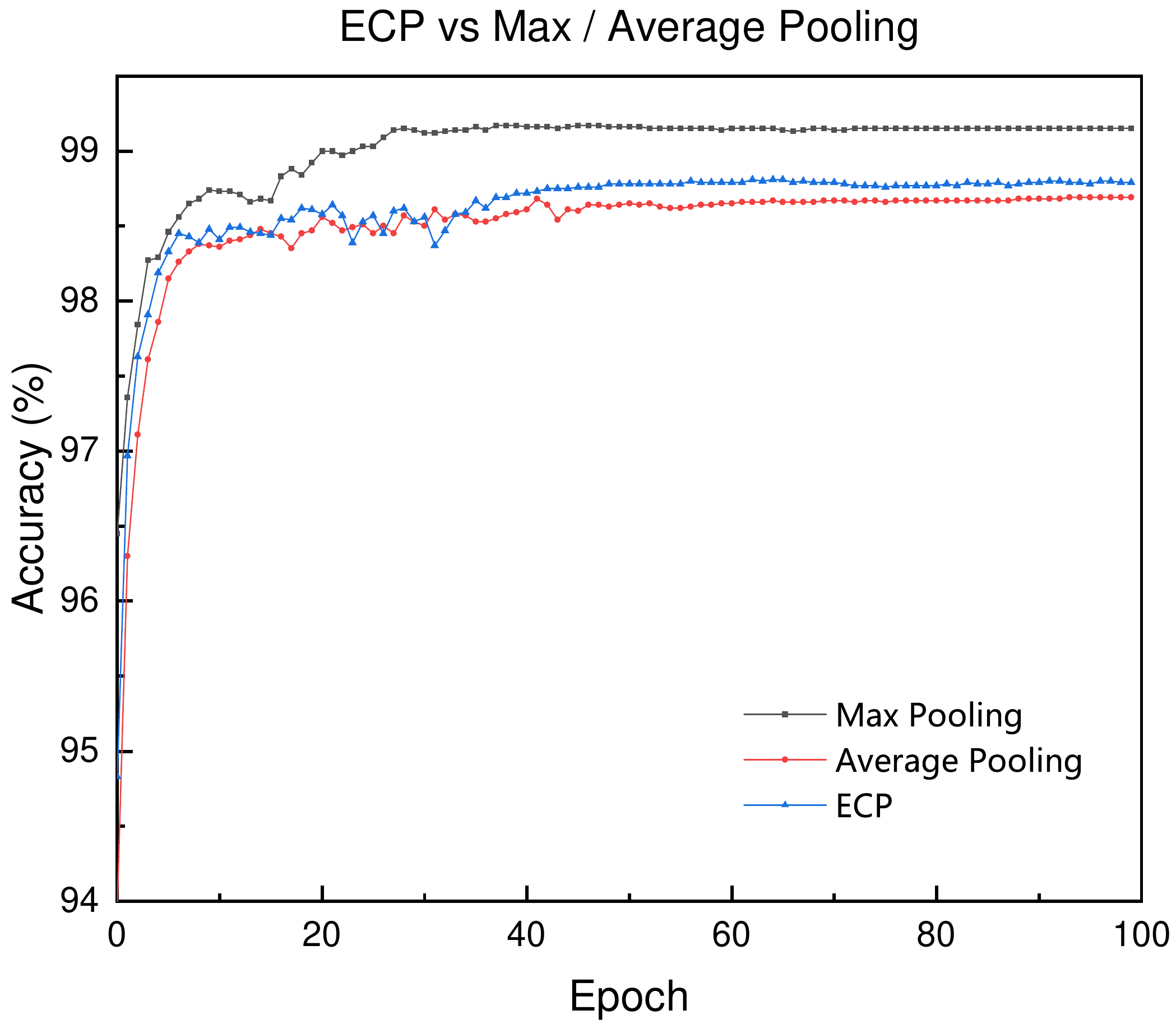}
\caption{ECP Convergence vs Average/ Max Pooling under Two-convolution Layer.}
\label{fig:10}
\end{figure}

Figure \ref{fig:7} and Figure \ref{fig:8} show Random Pooling convergence vs Average/ Max Pooling under one-convolution layer and two-convolution Layer respectively. Figure \ref{fig:9} and Figure \ref{fig:10} show ECP convergence vs Average/ Max Pooling under one-convolution layer and two-convolution Layer respectively.

Based on the experiments above, both Random Pooling and ECP can achieve good convergence compared to conventional Average/ Max Pooling.

\subsection{Remarks}
Based on the experiments above, in the following we summarize the major characteristics of the proposed ECP technique:
\begin{itemize}
\item Testing performance is always much better than training.
\item ECP has more advantage over Average Pooling than Max Pooling due to the speedup of training.
\item We can achieve more performance improvement when conducting ECP on a deeper network, with little loss in accuracy.
\end{itemize}

\section{Conclusions}
Deeper network architecture usually leads to better performance, as a result, it's getting more and more difficult to train Convolutional Neural Networks. Considering the fact that the overall execution time of Convolutional Neural Networks is dominated by convolution operations, we propose a novel technique named EasyConvPooling (ECP) to solve this problem. In ECP, we conduct convolution operations according to the index from following pooling layer, which reduces 75\% of original convolution operations. The experiments demonstrate that we achieve 1.45x speedup on training time and 1.64x on testing time with little loss in accuracy. What's more, we achieve a speedup of 5.09x on pure Easy Convolution operations compared to conventional convolution operations.

\bibliographystyle{ACM-Reference-Format}
\bibliography{sample-bibliography}

\end{document}